\documentclass[nohyperref]{article}

\usepackage[dvipsnames]{xcolor}         %

\usepackage{microtype}
\usepackage{graphicx}
\usepackage{subfigure}
\usepackage{booktabs} %

\usepackage{hyperref}

\usepackage[preprint]{icml2022}

\usepackage{amsmath}
\usepackage{amssymb}
\usepackage{mathtools}
\usepackage{amsthm}

\usepackage[capitalize]{cleveref}

\usepackage{amsmath, amssymb, amsthm, amsopn}       %
\usepackage{mathtools}
\usepackage{enumitem}
\usepackage{MnSymbol,wasysym}
\usepackage{graphicx}
\usepackage{empheq} %
\usepackage{cancel} %
\usepackage{mysymbols}      %
\usepackage{caption}

\icmltitlerunning{Model-Advantage \nmi{and Value-Aware Models} for Model-Based RL: \nmi{Bridging the Gap in Theory and Practice}}
\icmltitlerunning{}

\begin{document}

\twocolumn[
\icmltitle{
Model-Advantage \nmi{and Value-Aware Models} for Model-Based\\Reinforcement Learning: \nmi{Bridging the Gap in Theory and Practice}}

\icmlsetsymbol{equal}{*}

\begin{icmlauthorlist}
\icmlauthor{Nirbhay Modhe}{gt}
\icmlauthor{Harish Kamath}{gt}
\icmlauthor{Dhruv Batra}{gt}
\icmlauthor{Ashwin Kalyan}{aitwo}
\end{icmlauthorlist}

\icmlaffiliation{gt}{Georgia Institute of Technology}
\icmlaffiliation{aitwo}{Allen Institute for AI}

\icmlcorrespondingauthor{Nirbhay Modhe}{nirbhaym@gatech.edu}

\icmlkeywords{Machine Learning, ICML}

\vskip 0.3in
]

\printAffiliationsAndNotice{}  %

\begin{abstract}
    \nmi{
This work shows that value-aware model learning, known for its numerous theoretical benefits,
is also practically viable for solving challenging continuous control tasks in 
prevalent model-based reinforcement learning algorithms.
First, we derive a novel value-aware model learning objective by 
bounding the model-advantage i.e. model performance difference, between two MDPs or models
given a fixed policy, achieving superior performance to prior value-aware objectives in most
continuous control environments. 
Second, we identify the issue of stale value estimates in naively
substituting value-aware objectives in place of maximum-likelihood in dyna-style 
model-based RL algorithms. Our proposed remedy to this issue bridges the long-standing
gap in theory and practice of value-aware model learning by enabling successful deployment
of all value-aware objectives 
in solving several 
continuous control robotic manipulation and locomotion tasks. Our results are obtained
with minimal modifications to two popular and open-source model-based 
RL algorithms -- SLBO and MBPO, without tuning any existing hyper-parameters, while
also demonstrating better performance of value-aware objectives than these 
baseline in some environments.
}
\end{abstract}

\section{Introduction}
Reinforcement Learning (RL), with its many success stories \citep{mnih2015human, silver2016mastering, silver2017mastering, levine2016end, gu2016continuous}, has emerged as a promising learning paradigm.
These milestones are largely due to model-free RL approaches that come at a significant price in terms of sample efficiency. 
In fields like robotics or healthcare, obtaining large amounts of data is both impractical and expensive, making these methods ill-suited despite their successes in complex game environments.
\nm{As a result}, the alternate \nm{data-efficient} approach of Model-based Reinforcement Learning (MBRL), 
\nm{has become an increasingly important direction for the research community}.
\ak{However, the 
\nmi{status quo} 
of model learning is confined to mimicking real world data as opposed to learning representations that can induce optimal behavior; thereby reducing the scope of current MBRL approaches.} 

Traditional MBRL approaches seek to accurately learn the dynamics of the environment and in practice, employ maximum likelihood estimation (MLE) to achieve this -- \nm{\ie} minimizing the KL divergence between predicted and observed next state \nm{distributions}.
\nm{A drawback of this approach is the issue of an objective mismatch between the model-learning
objective and the ultimate purpose of using the model to find an optimal policy 
\citep{wang2019benchmarking, lambert2020objective}}.
More recent research in MBRL has focused on efforts to overcome these shortcomings -- including optimizing for auxiliary objectives \citep{lee2020context, nair2020goal, tomar2021model}, augmenting model-learning with exploration strategies \citep{janner2019trust, kidambi2020morel}, meta-learning to closely intertwine the two objectives \citep{nagabandi2018learning} and introducing inductive biases to the model-learning objective \citep{lu2020dynamics}.
However, these MBRL approaches still employ MLE.

In this work, we revisit Value Aware Model Learning (VAML) \citep{farahmand2017value, farahmand2018itervaml}, 
an alternate objective for learning dynamics.
Instead of predicting the exact next state, 
VAML seeks to predict states that have similar value as the observed next state.
For instance, suppose states $\sprime$ and $\sdoubleprime$ \nma{are distinct but} have the same value  
\ie $V^\pi(\sprime) = V^\pi(\sdoubleprime)$ under some policy $\pi$.
MLE-based model learning would penalize predicting $\sdoubleprime$ instead of the observed
state $\sprime$ since they are not identical. In VAML, $\sprime$ and $\sdoubleprime$ are
equivalent since they have the same value.
This objective is appealing as it factors in the \emph{utility} of the model in finding 
the optimal policy (through the value function) and does not require \nm{\emph{exact}}
prediction of observed trajectories.
\nmi{
Value-aware model-based RL has recently witnessed several theoretical advancements in the form
of guarantees of convergence \cite{farahmand2017value,farahmand2018itervaml}, 
the value-equivalence principle \cite{grimm2020value}
and use optimistic model-based RL for regret minimization \cite{ayoub2020model}.
The MuZero algorithm \citep{muzero2020} is also an example of a value-aware (or `value-equivalent')
model-based approach for solving discrete action environments while leveraging Monte-Carlo tree search.
}

Despite the intuitive and theoretical appeal of \nm{existing} value-aware model 
learning objectives, their utility 
has thus far remained under-explored beyond toy 
\ak{settings 
}
\nmi{for continuous control}.
\nm{In our experiments, \hk{we} found that existing value-aware objectives 
perform poorly with model-based RL frameworks, 
independently replicating recent
negative results \citep{lovatto2020decision}}.
\nmi{
In this work, we first derive an upper bound on the expected model
performance difference of two MDPs or models for a fixed policy, using
triangle inequality on the $L1$-norm.
In contrast, prior value-aware approaches \citep{farahmand2018itervaml},
though inspired by the minimization of (normed) model performance difference,
do not upper bound the model performance difference with their use of the $L2$-norm,
which may explain their inferior performance compared to our proposed 
objective in most of our continuous control experiments.
}
\nmi{
Secondly, we discover the key issue of stale value estimates in
the naive application of value-aware losses in the 
dyna-style model-based RL algorithmic framework.
Upon correcting for the stale value estimates by intermittently fitting the value
network during model learning, we obtain significant performance
improvements on the more challenging continuous control environments.}
\nmi{The resulting general purpose dyna-style MBRL algorithm is,}
to the best of our knowledge,
the first known practical deployment of value-aware objectives 
in challenging continuous control domains including MuJoCo 
\citep{todorov2012mujoco} robotic simulation environments \citep{brockman2016openai}.

\nmi{
We empirically test our proposed algorithm and novel upper bound on two
recent dyna-style MBRL algorithms -- SLBO \cite{luo2018algorithmic} and MBPO \cite{janner2019trust}.
We find that our algorithm, without tuning \emph{any existing hyperparameter}, 
successfully bridges the gap in theory and practice by reaching near-matching
performance \wrt MLE-based baselines in most continuous control simulation tasks and outperforming
them in some others.
}
\nm{We hope that these encouraging results spur wider interest in the community leading to both 
adoption and further study of \ak{value-aware methods for 
\nmi{practical model-based RL.}}
}

\section{Related Work}
\textbf{MLE-based MBRL.}
\nma{Maximum likelihood estimation (MLE)}
is the most prevalent and straight-forward objective for model learning
\nma{in an MBRL framework \citep{sutton1990dyna,sutton2012dyna}}, with the goal of
modeling state transitions accurately.
Unlike our proposed value-aware objective, minimizing 
\nmi{MLE}
error minimizes a looser upper bound on the model performance difference \nma{\cite{farahmand2017value}}.
Therefore, multiple MBRL approaches that minimize various definitions of dynamics error have been proposed.
For instance, \citet{azar2012sample, azar2013minimax, azar2017minimax} use \naive empirical frequencies.
More sophisticated approaches use function approximators and minimize various statistical distances -- \eg KL \cite{ross2012agnostic},  total-variation \cite{janner2019trust} or Wasserstein metrics \cite{wu2019model}.

\textbf{Non-MLE based MBRL.} 
\nma{Methods that inform model learning via the value function, reward or policy have recently gained popularity \citep{oh2017value,silver2017predictron,muzero2020,abachi2020policy}}.
\nma{In particular,} \citet{muesli}, \citet{muzero2020}, and \citet{muzero2021} 
explore learning dynamics implicitly using the estimated value for a given state, and using a \nma{Monte-Carlo} tree search algorithm to plan with this learned model -- their method, MuZero, is an instance of a value-equivalent \cite{grimm2020value} model learning approach.
However, these works learn a joint model for directly estimating future values and actions (policy) \nma{without any explicit future predictions in the state space}.
\nma{In contrast, we focus on the class of MBRL methods that explicitly make predictions in the state space, allowing for simple adaptations on top of of well-known MBRL frameworks e.g. Dyna-style algorithms \citep{sutton1990dyna}} such as SLBO \cite{luo2018algorithmic} and MBPO \cite{janner2019trust}.

\textbf{Value-aware Model Learning.}
\citet{farahmand2017value}, \citet{farahmand2018itervaml}, \citet{ayoub2020model} and \citet{grimm2020value} are the closest prior works that study the theoretical properties of value-aware objectives,
\nmi{with experiments restricted to pedagogical settings with small state spaces or the cart-pole environment.}
\nma{
\citet{lovatto2020decision} demonstrate \textit{negative} empirical results for their practical instantiation of value aware model learning \citep{farahmand2018itervaml} with an actor-critic learner in continuous control environments such as \texttt{Pusher-v2} and \texttt{InvertedPendulum-v2}. Their algorithm employs a sparse model update, occurring only once every few policy and critic updates -- different from our algorithm that builds on top of a standard Dyna-style MBRL algorithm with multiple model updates in between every sequence of agent updates. 
}

\section{Preliminaries}

\subsection{Markov Decision Processes}
In this work, we consider a discrete-time infinite-horizon RL problem characterized by Markov Decision Processes (MDPs) $M$ defined as $(\calS, \calA, \calP, \calR, \calP_0, \gamma)$.
Here, $\calS$ is the state space, $\calA$, the action space, $\calP:\calS\times\calA\times\calS \rightarrow \mathbb{R}$ the transition probabilities or \emph{dynamics}, $\calR:\calS\times\calA \rightarrow [0,R_{max}]$ the reward function, $\calP_0$ the starting state distribution and finally, $\gamma$ the discount factor.
The goal is to find the \emph{optimal} policy $\pistar\in\Pi$ that maximizes the (discounted) total return $J:\pi\rightarrow\mathbb{R}$ \ie $J(\pi) = \EE_{\rho^{\pi}}\left[\sum_t \calR(s_t, a_t)\right]$.
where $\rho^\pi$ is the distribution of trajectories $(s_0,a_0,s_1, \dots)$, $s_0\sim\calP_0$, when acting according to policy $\pi$.

The Q-function and the value function under policy $\pi$ are given by
$Q^\pi(s,a) = \EE_{\rho^{\pi}}\left[\sum_t\calR(s_t,a_t) \mid \pi, s_0=s,a_0=a \right]$ and $V^\pi(s) = \EE_{\rho^{\pi}}\left[\sum_t \calR(s_t,a_t) \mid \pi, s_0=s\right]$.
A more useful version of the value function, and therefore the RL objective itself, is obtained by defining the \emph{future state distribution} $P^\pi_{s,t}(s') = \Pr(s_t=s'\mid \pi, s_0=s)$ and $\gamma$-discounted 
stationary state distribution
$d_{s,\pi}(s') = (1-\gamma) \sum_{t=0}^\infty \gamma^t P^\pi_{s,t}(s')$, 
where we drop the dependency on start state distribution when it is 
implicitly assumed to be known and fixed.
Using these definitions, we write the value function as:
\begin{equation}\label{eq: useful_value}
\begin{aligned}
    V^\pi(s_0) &= \sum_{t=0}^\infty\gamma^t\EE_{(a_t,s_t)\sim(\pi,P^{\pi}_{s_{0},t})}[\calR(s_t,a_t)] \\
    &= \frac{1}{1-\gamma}\EE_{a,s\sim \pi,d_{s_0,\pi}} [\calR(s,a)]
\end{aligned}
\end{equation}

\begin{algorithm}[tb]
\caption{Model Based Reinforcement Learning (MBRL)}
\label{alg:mbrl}
\begin{algorithmic}
    \STATE Randomly initialize policy $\pi$, model $\rM$
    \STATE Initialize replay buffer $\calD\leftarrow\emptyset$
    \FOR {$n_{outer}$ iterations}
        \STATE // \textit{model update step}
        \FOR{$K_{model}$ updates }
            \STATE $\calD \leftarrow \calD \bigcup $\{$n$ samples from true environment $\bM$ collected by $\pi$\}
            \STATE Update $\bM$ using model-learning objective on $\calD$ // \textit{\eg $\mathbb{E}_\calD\left[KL(\hat{s}||s)\right]$} \label{line:model_update}
        \ENDFOR
    \ENDFOR
    \STATE // \textit{policy update step}
    \FOR{$K_{policy}$ updates }
        \STATE $\calD^{\prime} \leftarrow $ \{Samples collected in learned model $\rM$ using $\pi$.\}
        \STATE Update $\pi$ using policy learning method // \textit{\eg TRPO \citep{schulman2015trust}}
    \ENDFOR
\end{algorithmic}
\end{algorithm}

\subsection{Model-Advantage and MBRL} 
MBRL algorithms work by iteratively learning an 
approximate model and 
\nm{then deriving an optimal policy from this model either by planning 
\nmi{with MPC (model-predictive control)} or 
learning a separate policy \nmi{(actor-critic)} with imagined experience.
The latter case refers to the family of Dyna-style MBRL algorithms \citep{sutton1990dyna} that we adopt
in this work -- see \cref{alg:mbrl} for a representative algorithm from this family.}
Model-advantage\footnote{
    Name follows policy-advantage that compares the utility of two actions \citep{kakade2002approximately}
},
proposed by \nma{\cite{metelli2018configurable, modhe2020bridging}}, is a key quantity that can be used to compare the utility of transitioning according to the approximate model $\rM$ as opposed to the true model $\bM$.
Specifically, \emph{model-advantage}
denoted by $\MA^\pi_{\rM}(s,s')$ compares the utility of moving to state $s'$ and thereafter following 
the trajectory governed by model $\rM$ as opposed to following $\rM$ from 
state $s$ itself; while acting according to policy $\pi$.
The following definition in \cref{eq: model-advantage} captures this intuition.
We denote model-dependent quantities with the model as subscript: 
transition probability distribution of $\rM$ is denoted by $P_{\rM}$ and value function
as $V^{\pi}_{\rM}$.
\begin{equation}\label{eq: model-advantage}
\MA^\pi_{\rM}(s,s') := 
    \gamma\left[
        V^\pi_{\rM}(s') 
        - \EE_{s''\sim P_{\rM}(s,\pi)}
            V^\pi_{\rM}(s'')
    \right]
\end{equation}
Here, $V^\pi_{\rM}$ is the model-dependent value function defined as:
\[V^\pi_{\rM}(s) = 
    \EE_{\rho^\pi_{\rM}}
    \left[
        \sum_{t=0}^\infty \gamma^{t} \calR_{\rM}(s_t,a_t) \mid \pi, \rM, s_0 = s 
    \right]\]
We are now ready to restate the well-known \emph{simulation lemma} \citep{kearns2002near} that quantifies the model performance difference using model-advantage.
\begin{lemma}\label{lemma: model-perf-diff}
    (Simulation Lemma) Let $\rM$ and $\bM$ be two different MDPs.
    Further, define 
    $\calR^\pi_{M}(s) = \EE_{a\sim\pi(\cdot|s)}[\calR_{M}(s,a)]$ 
    and $\calR^{\pi}_{\delta_{\rM,\bM}}(s) = \calR^\pi_{\rM}(s) - \calR_{\bM}^\pi(s)$.
    For a policy $\pi\in\Pi$ we have:
    \begin{equation}\label{eqn: model-perf-diff}
    \begin{aligned}
        J_\rM(\pi)
        &= J_\bM(\pi) 
        + \underbrace{
        \EE_{s\sim d_{\rM,\pi}}[\calR^{\pi}_{\delta_{\rM,\bM}}(s)]
        }_{\text{\textup{reward difference}}}\\
        &+ \frac{1}{1-\gamma} 
            \underbrace{
                \EE_{s\sim d_{\rM, \pi}}
                \EE_{s'\sim \Ptrue_{\rM}(s,\pi)}
                    \left[
                        \MA^{\pi}_{\bM}(s,s')
                    \right] 
            }_{\text{\textup{expected model-advantage}}}\\
    \end{aligned}
    \end{equation}
\end{lemma}
Here, we use a model-dependent stationary state distribution 
$d_{\rM,\pi}(s)
$ (dropping the dependence on start state distribution) 
where the dynamics $\calP_{\rM}$ are used.
To simplify notation, we will write the expected model advantage term as
$\EE_{(s,s')\sim\rM}\left[\MA^{\pi}_\bM(s,s')\right]$ or simply 
\EMA.
A slightly different form of \cref{lemma: model-perf-diff} can be obtained by explicitly indicating the model in the Bellman operator as follows.
\begin{align}
\calT^\pi_{\rM} V(s) := 
    \EE_{a\sim\pi}
    \left[
        \calR_{\rM}(s,a) 
        + \gamma \EE_{s'\sim P_{\rM}(s,a)}[V(s')]
    \right]
\end{align}
This leads to the following corollary that provides an alternate view of the model-advantage term (see 
\nma{Appendix}
for the proof).
\begin{corollary}\label{cor: model-perf-diff}
    Let $\rM$ and $\bM$ be two different MDPs.
    For any policy $\pi\in\Pi$ we have:
    \begin{align}
        &J_\rM(\pi) = 
        J_\bM(\pi) 
        \nonumber\\&\phantom{AAAA}
        + \frac{1}{1-\gamma}\EE_{s\sim d_{\rM, \pi}}
        \underbrace{
            \left[
                \calT^\pi_{\rM} V_{\bM}^\pi(s) 
                - \calT^\pi_{\bM} V_{\bM}^\pi(s)
            \right]
        }_{\text{\textup{deviation error}}} 
        \label{eq: dev}
    \end{align}
\end{corollary}
Note that the term on the right that includes the deviation error 
is exactly equal to model-advantage when the reward functions 
of the two MDPs are identical\footnote{A common assumption 
for MBRL works proposing to learn dynamics (\eg \cite{luo2018algorithmic}). 
We make this assumption as well.
}.
Therefore, setting aside the reward-error term in \cref{lemma: model-perf-diff}, model advantage can be viewed as the deviation resulting from acting according to different MDPs. 
Minimizing the deviation error is the basis of the objective proposed in Value-Aware Model Learning (VAML) \cite{farahmand2017value, farahmand2018itervaml}.
More recent work \citep{grimm2020value} shows that various MBRL methods can be thought of as minimizing the deviation error -- a direct consequence of the close relationship between the deviation error and the 
\nm{model performance difference}.

\section{Approach: Model Advantage Based Objective}
\colorM

In this section, we first introduce the basis of value-aware model learning 
where the objective is to minimize the
performance difference of a policy in the true vs approximate model.
From Eqn.~\ref{eqn: model-perf-diff}, this translates to optimization of
expected model advantage \EMA, for which we show an empirical
estimation strategy with samples from the true MDP and gradient based
updates for a parametrized dynamics model.
We then derive a novel upper bound on expected model advantage 
and introduce a general purpose algorithm for value-aware model-based RL.

\subsection{Optimizing Model Advantage}
For the model-learning step in MBRL, we are interested in an objective for finding 
model parameters $\rphi$ corresponding to the dynamics of the approximate MDP 
\ie $\calP_{\rphi}(\cdot\mid s,a)$ that eventually lead to the learning of an 
optimal policy in the true MDP $\trueM$.
By looking at the model-advantage version of the simulation lemma 
(\ie \cref{lemma: model-perf-diff}), a natural choice for a loss function is the 
absolute value of the expected model advantage.
\nmi{For brevity, we replace the expectation over ${s_t\sim P^\pi_{\trueM,t}, s_{t+1}\sim P^\pi_{\trueM,t+1}}$
with $\tilde{d}_{t}$.}
\begin{align}
&\calL_1(\clr{\phi}) := 
\left| 
    J_\rMphi(\pi) - J_\trueM(\pi) 
\right|
\notag
\\
&=
\Bigg|
\sum_{t=0}^\infty \gamma^t  \mathop{\EE}\limits_{
    \nmi{\tilde{d}_{t}}
}
\bigg[
    \gamma V^\pi_{\rMphi} (s_{t+1})
    - \gamma \mathop{\EE}\limits_{\sdoubleprime\sim \calP^\pi_\rMphi(s_t, \pi)}
        \left[
            V^\pi_\rMphi(\sdoubleprime)
        \right]
\bigg]
\Bigg|
\label{eqn: direct_abs_ma}
\end{align}
This objective can be empirically estimated via 
\nmi{
trajectories $ (s_0, a_0, \dots, a_{T-1}, s_T) \in \calD_{m}$
where dataset $\calD_{m}$ is
}
sampled from the true MDP $\trueM$. We omit the input of $\pi$ in $\calP^\pi_{\rMphi}$.
\begin{align}
&\widehat{\calL_1}(\clr{\phi})
=\notag\\
&\Bigg|
\sum_{
    \calD_{m}
}
\sum_{t = 0}^{T-1}
\frac{\gamma^{t}}{m}
\gamma^{t}
\bigg(
    V^\pi_{\rMphi} (s_{t+1})
    - 
    \mathop{\EE}\limits_{\sprime \sim \calP^{\pi}_{\rMphi}(s_{t})}
    \left[
        V^\pi_\rMphi(\sprime)
    \right]
\bigg)
\Bigg|
\label{eqn: direct_abs_ma_empirical_intermediate}
\end{align}
In Eqn.~\ref{eqn: direct_abs_ma_empirical_intermediate}, the value function $V^{\pi}_{\rMphi}$
has a complex dependency on parameters $\rphi$ which is \nmi{difficult} to optimize. In practice,
\nmi{as observed by \citet{farahmand2018itervaml},}
this value can be estimated in any Dyna-style \citep{sutton1990dyna} model-based RL algorithm
with a parametrized value function 
(with parameters included in $\theta$ of policy $\pi_{\theta}$ \nma{i.e. 
as part of an actor-critic pair}) 
for estimating this value.
We estimate the value function without modeling its dependency on $\rphi$
\nma{i.e. we replace $V_{\rMphi}^{\pi}$ with a learned value network 
$V^{\pi_{\theta}}$, such that the
$V^{\pi_{\theta}}$ is updated during the policy-update step of our algorithm
(using imagined experience from $\rMphi$)
to match the true target $V_{\rMphi}^{\pi}$.}
\nma{This results in} a simple stochastic gradient update rule\footnote{
Note that this objective can be optimized via gradient updates as long as the 
value function $V^{\pi_{\theta}}$ can be differentiated \wrt its inputs \ie states, \nma{which is the case for neural networks.}}
for 
\nmi{reparametrized samples from }
$P^{\pi}_{\rMphi}(s_{t})$ \nmi{(typically Gaussian)}. 
Finally, our empirical objective is as follows.

\begin{align}
&\widehat{\calL_1}(\clr{\phi})
=\notag\\&
\Bigg|
\sum_{
    \calD_{m}
}
\sum_{t = 0}^{T-1}
\frac{\gamma^{t}}{m}
\bigg(
    V^{\pi_{\theta}} (s_{t+1})
    - 
    \mathop{\EE}\limits_{\sprime \sim P^{\pi}_{\rMphi}(s_{t})}
    \left[
        V^{\pi_{\theta}}(\sprime)
    \right]
\bigg)
\Bigg|
\label{eqn: direct_abs_ma_empirical}
\end{align}

\begin{algorithm}[t]
\caption{Value-Aware MBRL}
\label{alg:valmbrl}
\begin{algorithmic}
\STATE Initialize $\theta=(\theta_p,\theta_v)$, the policy/value function parameters and model parameters $\rphi$ randomly\;
\STATE Initialize replay buffer $\calD, \calD^{\prime}, \calD^{\prime\prime} \leftarrow\emptyset$\;
\FOR{$K$ iterations}
    \STATE $\calD \leftarrow \calD \bigcup $\{$n$ samples from true environment $\trueM$ collected by $\pi_{\theta_p}$\}\;
    \STATE // \textit{model update step}
    \FOR{$K_{\text{model}}$ updates }
        \STATE Update $\rphi$ using value-aware model-learning objective on $\calD$; //\textit{\eg \cref{eqn: ma_l1_upper_bound_empirical}}
        \IF{every $K_{\text{interval}}$ model updates}
            \STATE // \textit{update stale value parameters}
            \STATE $\calD^{\prime\prime} \leftarrow $ \{$m$ samples collected in learned model $\rM$ using $\pi_{\theta_p}$\}\label{line:mod_val_start}\;
            \STATE Update $\theta_v$ to estimate discounted return with $D^{\prime\prime}$\label{line:mod_val_end}\;
        \ENDIF
    \ENDFOR \label{line:value_model_update}
\ENDFOR
\STATE // \textit{policy update step}
\FOR{$K_{\text{policy}}$ updates }
    \STATE $\calD^{\prime} \leftarrow $ \{$m$ samples collected with $\pi_{\theta_p}$ in model $\rM$\}\;
    \STATE Update $\theta_p, \theta_v$ using policy learning method //\textit{\eg TRPO \citep{schulman2015trust}}
\ENDFOR
\end{algorithmic}
\end{algorithm}

\subsection{Model-Advantage Upper Bound}
In practice, the objective in Eqn.~\ref{eqn: direct_abs_ma_empirical} is undesirable 
as it requires full length trajectory samples to compute the discounted sum and 
therefore, provides a sparse learning signal \ie a single gradient update step 
from an entire trajectory.
This limitation is overcome by further upper bounding \cref{eqn: direct_abs_ma} via the triangle inequality as shown below (with abbreviated notation).
\begin{align}
\calL_1(\clr{\phi})
:&=\bigg|
    \sum_{t=0}^\infty \gamma^t  \mathop{\EE}\limits_{\trueM,t}
    \left[
        \MA^{\pi}_\rM
    \right]
\bigg|
\leq
\underbrace{
    \sum_{t=0}^\infty \gamma^t  \mathop{\EE}\limits_{\trueM,t}
    \left[
    \left|
        \MA^{\pi}_\rM
    \right|
    \right]
}_{
=:
\calL^{U}_1(\clr{\phi})
}
\label{eqn: ma_l1_upper_bound}
\end{align}
Observe that this form of the objective is now compatable with experiences \ie $(s,a,s',r)$ sampled from the true MDP $\trueM$ as opposed to ensure trajectories -- thereby providing a denser learning signal.
We further make this objective amenable to minibatch training by replacing the the discounted sum over timesteps $\sum_{t=0}^{\infty} \gamma^t \EE_{s_t} (\cdot)$ with the policy's discounted stationary state distribution $\EE_{s\sim \rho_{\pi,\trueM}} (\cdot)$ -- 
this is estimated empirically with a finite dataset of sampled experiences.
Similar to \cref{eqn: direct_abs_ma_empirical}, the empirical estimation version of the objective is as follows,
\nmi{where the summation is over trajectories $(s_{t}, a_{t}, s_{t+1}) \in \calD_{n}$.}
\begin{align}
\widehat{\calL^{U}_1}(\clr{\phi})
&=
\frac{1}{n}
\sum_{
    \nmi{\calD_{n}}
}
\bigg|
    V^{\pi_{\theta}} (s_{t+1})
    - 
    \mathop{\EE}\limits_{\sprime \sim p^{\pi}_{\rMphi}(s_{t})}
    \left[ 
        V^{\pi_{\theta}}(\sprime)
    \right]
\bigg|
\label{eqn: ma_l1_upper_bound_empirical}
\end{align}

\nmi{In \Cref{sec: discrete_envs} we observe the benefits of the denser learning signal
provided by this upper bound in contrast with \Cref{eqn: direct_abs_ma_empirical}
on discrete environments where the both objectives converge successfully.
}

\textbf{Connection to VAML.} 
Eqn.~\ref{eqn: ma_l1_upper_bound} is similar to the L2 norm value-aware objective introduced in \cite{farahmand2017value,farahmand2018itervaml}.
In our framework, the VAML objective, $\lvaml$, can be obtained by using the L2 norm in \cref{eqn: ma_l1_upper_bound} \ie $\lvaml:=\sum_{t=0}^\infty \gamma^t  \mathop{\EE}_{\trueM,t}
    \left[
    \left(
        \MA^{\pi}_\rM
    \right)^2
    \right]$.
Importantly, owing to the properties of L2 norm, note that $\lvaml$ does \emph{not} upper bound its corresponding L2 normed model advantage objective $\calL_{2}$.
We find in our experiments that $\calL^{U}_{1}$ has better overall 
performance \nmi{in conjunction with SLBO}, 
potentially hinting at the importance of this relationship 
with model-advantage. 

\begin{figure*}[t!]
\centering
    \resizebox{0.86\textwidth}{!}{\includegraphics[width=1.0\textwidth]{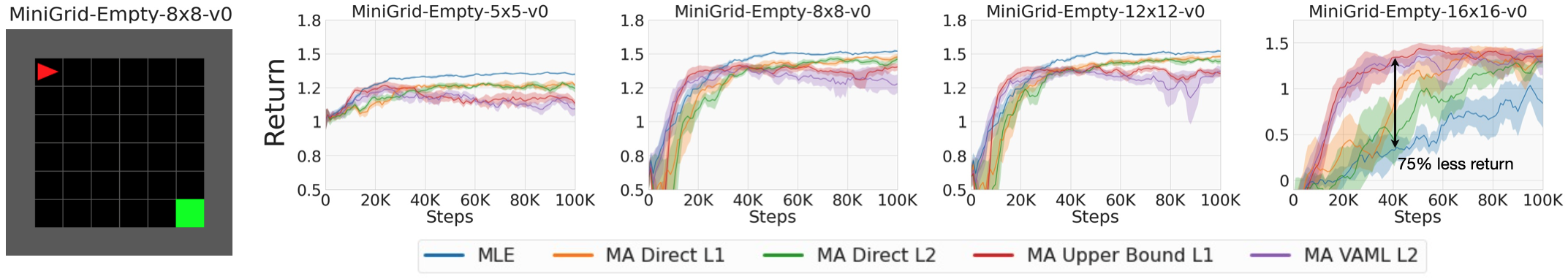}
    }
    \caption{(Left) 
    A sample of the \texttt{8x8} size gridworld environment
    from Gym MiniGrid 
    \protect\citep{gym_minigrid}. (Right) Return curves over 5 random seeds 
    on \nmi{\texttt{MiniGrid-Empty} environments
    with grid sizes \texttt{5x5}, \texttt{8x8}, \texttt{12x12} and \texttt{16x16} 
    }
    using 4 value aware methods and an MLE baseline.
    Increasing grid size \nmi{(up to \texttt{16x16})} negatively affects MLE performance most, while 
    our proposed upper bound and 
    VAML~\protect\citep{farahmand2018itervaml} are affected the least.
    The direct versions of L1 and L2 model-advantage based objectives (MA Direct L1 and MA Direct L2)
    are further slower to converge than MA Upper Bound L1 and MA VAML L2.
    }
    \label{fig:minigrid}
\end{figure*}

\subsection{General Algorithm for Value-aware Objectives}

Value-aware objectives such as \cite{farahmand2018itervaml, grimm2020value} enjoy several
theoretical benefits, but remain isolated from practical use 
beyond small, finite state toy MDPs.
We find that with
a naive substitution of value-aware objectives in place 
of maximum likelihood (\Cref{fig:ablation_naive}) 
in existing model-based RL algorithms 
(\ie \cref{line:model_update} in \cref{alg:valmbrl}) 
\nmi{
worked well only for the easy continuous control environments,
namely \texttt{Cartpole}, \texttt{Pendulum} and \texttt{Acrobot}.
}
\nmi{
This supports the evidence in
\citet{lovatto2020decision} where negative results were demonstrated for 
their choice of simple
environments -- \texttt{Pusher-v2} and \texttt{InvertedPendulum-v2}, and 
value-aware errors alone were examined 
for \texttt{Hopper-v3} and \texttt{Walker2d-v3}.
}
\nmi{
Next, we describe our algorithm with which 
we find positive results in conjunction with the SLBO algorithm \cite{luo2018algorithmic}
on \texttt{Swimmer-v1}, \texttt{Hopper-v1} and \texttt{Ant-v1} environments
and in conjunction with the MBPO algorithm \cite{janner2019trust} on 
\texttt{Walker-v2} and \texttt{Hopper-v2} environments.
}

\subsubsection{Correcting Stale Value Estimation}

\nmi{\Cref{alg:mbrl} represents the standard
framework of a Dyna-style algorithm,}
where the model is trained in a model-update step with 
\emph{ground truth} experience (samples from $\trueM$) 
and the policy and value parameters 
$\pi_{\theta}, V_{\theta} := V^{\pi_{\theta}}$ 
are trained in a policy update step with 
\emph{virtual} \nma{or imagined experiences} (samples from $\rMphi$).

\nmi{
Value-aware objectives have an additional dependency on value estimates in model learning in 
the from of $V^{\pi_{\theta}}$ 
(\cref{eqn: direct_abs_ma_empirical}) which play the role of $V^{\pi}_{\rMphi}$
in simplifying \cref{eqn: direct_abs_ma_empirical_intermediate}.
However,} 
dropping the dependency on $\rphi$  in $V^{\pi}_{\rMphi}$ from
Eq. \ref{eqn: direct_abs_ma_empirical_intermediate} to \ref{eqn: direct_abs_ma_empirical}
leads to an issue of stale value estimates in the default 
dyna-style algorithm, described as follows.
For every model update, the parameter 
$\rphi$ of $\rMphi$ is changed and as a result, 
the value function term in \cref{eqn: direct_abs_ma_empirical_intermediate}
\nmi{no longer corresponds to the same $\rMphi$.}
This implies that for multiple consecutive model updates 
with a fixed $V^{\pi_{\theta}}$,
the target that $V^{\pi_{\theta}}$ is supposed to estimate 
has moved -- making it a stale estimate.

In \Cref{alg:valmbrl} 
\nmi{we remedy this issue by updating}
the value function 
intermittently (while keeping policy fixed) between 
model updates. 
\nma{
Such an intermittent update is relatively cheap to perform as 
(i) it does not rely on any additional ground truth experience, 
(ii) it updates solely the value network and not the policy network, and 
(iii) the frequency of intermittent updates need not be very high \nmi{-- 
controlled by the new hyperparameter \nmi{$K_{\text{interval}}$} in \Cref{alg:valmbrl}. 
We found that setting $K_{\text{interval}}$ to $20$ for SLBO and $5$ for MBPO works well in practice.}
Intuitively, such intermittent value updates allow for a 
novel interplay in the form of a joint optimisation of 
model estimates and value estimates \nmi{(keeping policy fixed)} in conjunction with 
any value aware model learning objective. We hypothesize that this interplay
adds stability to the optimisation of value aware objectives 
\nmi{and in the next section, 
we verify it's role in the same with an ablation experiment (\cref{fig:ablation_naive}).}
}

\section{Experiments}

\begin{figure*}[t!]
\centering
    \resizebox{0.9\textwidth}{!}{
        \includegraphics[width=0.98\textwidth]{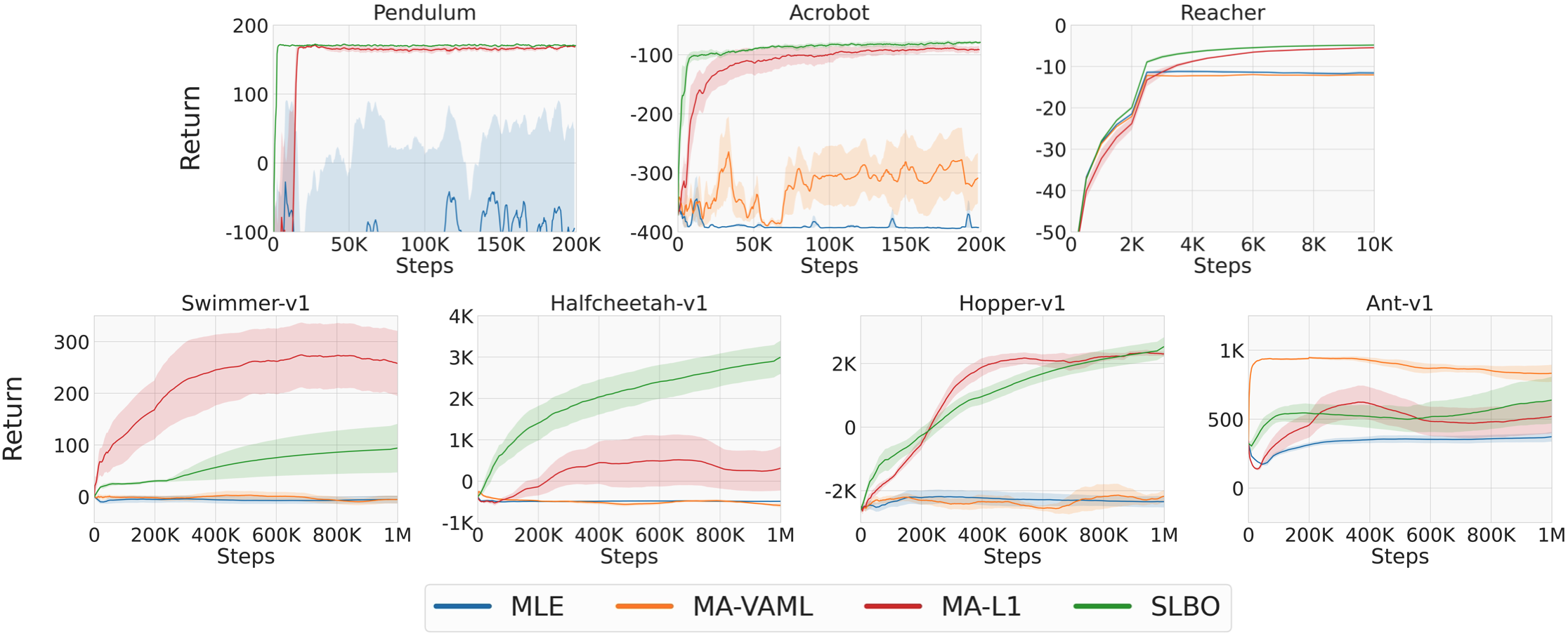}
    }
\caption{%
Evaluation on continuous control environments 
for value aware methods and baselines
with SLBO~\protect\citep{luo2018algorithmic},
\nmi{ without tuning existing parameters, over 5 seeds}.
Our objective \texttt{MA-L1} 
achieves better return and sample efficiency in comparison
to \texttt{MA-VAML} on most environments (\Ant{} being the exception)
\nmi{and in comparison to \texttt{MLE} on all environments.}
On the \Swimmer{} and \Hopper{} environments,
we also outperform \nma{or are competitive with} \texttt{SLBO}.
}
\label{fig:continuous_control_slbo}
\end{figure*}

In this section, we investigate our model learning objective in the context
of model based reinforcement learning in \nm{two} settings. 
First, we evaluate our algorithm in a discrete-state MDP where we 
optimize expected model advantage directly or indirectly via 
Eqns.~\ref{eqn: direct_abs_ma_empirical}~,~\ref{eqn: ma_l1_upper_bound_empirical},
\nmi{with the purpose of establishing the performance and convergence 
relationship among the selected value-aware
objectives and a maximum-likelihood baseline in a pedagogical setting.}
\nmi{Second, we evaluate \Cref{alg:valmbrl} together with
our proposed and a prior value aware objective
on several continuous control tasks, with two recent
dyna-style MBRL algorithms --
SLBO \citep{luo2018algorithmic} \nmi{and MBPO \citep{janner2019trust}}.
}

\subsection{Discrete \nmi{State and Control}}
\label{sec: discrete_envs}

We first establish the efficacy of value-aware objectives in a finite state 
setting with increasing state space size.
For this experiment, we use a discrete-state episodic gridworld MDP with 
cardinal actions \texttt{\{North, South, East, West\}}, an $N\times N$ grid,
deterministic transitions and
a fixed, absorbing goal state located at the bottom right of the grid
and agent spawning at the top left. A dense reward is provided for improvement
in L2 distance to the goal square and an additional decaying reward is provided upon
reaching the goal. The environment is empty except for walls along all edges.
\nmi{
Since the values of the optimal
policy are symmetric around the major diagonal of the grid it should provide
a slight advantage for value-aware methods that learn the value-equivalence of 
such states.}
We use 
\nmi{four configurations} of grid sizes: 
\nmi{$5\times 5$, $8\times 8$, $12\times 12$ and $16\times 16$}.
\nmi{
The dynamics model's prediction space for the next state is a discrete
set of the total number of states or grid cells in each environment --
the increasing grid size quadratically increases the total number of states
and hence, makes model learning challenging.
}

\textbf{Methods.} We denote \texttt{MA Direct L1} and \texttt{MA Direct L2}
as methods that optimize $\calL_{1}$ and $\calL_{2}$ objectives 
respectively 
(Eq. \ref{eqn: direct_abs_ma_empirical}).
\texttt{MA Upper Bound L1}
optimizes our proposed upper bound $\calL^{U}_1$ and 
\texttt{MA VAML L2} optimizes
$\lvaml$ (IterVAML from \citep{farahmand2017value}).
In computing the objectives from equations 
\ref{eqn: direct_abs_ma_empirical} and \ref{eqn: ma_l1_upper_bound_empirical},
the expectation over
predicted states is computed exactly as a summation over all states. 
\texttt{MLE} denotes the maximum likelihood baseline. 
For all methods, we use 
A2C as the policy update protocol in the MBRL algorithm.

\textbf{Results.} Figure \ref{fig:minigrid} shows return curves for 
all methods and environment configurations. Return greater than 1 corresponds
to reaching the goal (green square) and solving the task successfully and higher
returns correspond to fewer steps taken to reach the goal.
We observe that \texttt{MLE} 
sample efficiency decreases with increase
in grid size (left to right in \cref{fig:minigrid}) 
and all value based methods outperform this baseline 
\nmi{on the largest grid size
of \texttt{16x16}}. 
We observe that
the upper bounds on expected model advantage 
\texttt{MA Upper Bound L1} and \texttt{MA Upper Bound L2} 
achieve better sample efficiency
than the direct counterparts \texttt{MA L1} and \texttt{MA L2},
which is expected due to the sparser learning signal from the norm
of the summation over value differences in the direct computation 
as opposed to sum of normed value differences in the upper bounds.
\nmi{In conclusion, we find that value-aware methods do outperform
maximum-likelihood in discrete state settings with increasing number of states.
}

\begin{figure*}[t]
\centering
    \resizebox{0.89\textwidth}{!}{
        \includegraphics[width=0.98\textwidth]{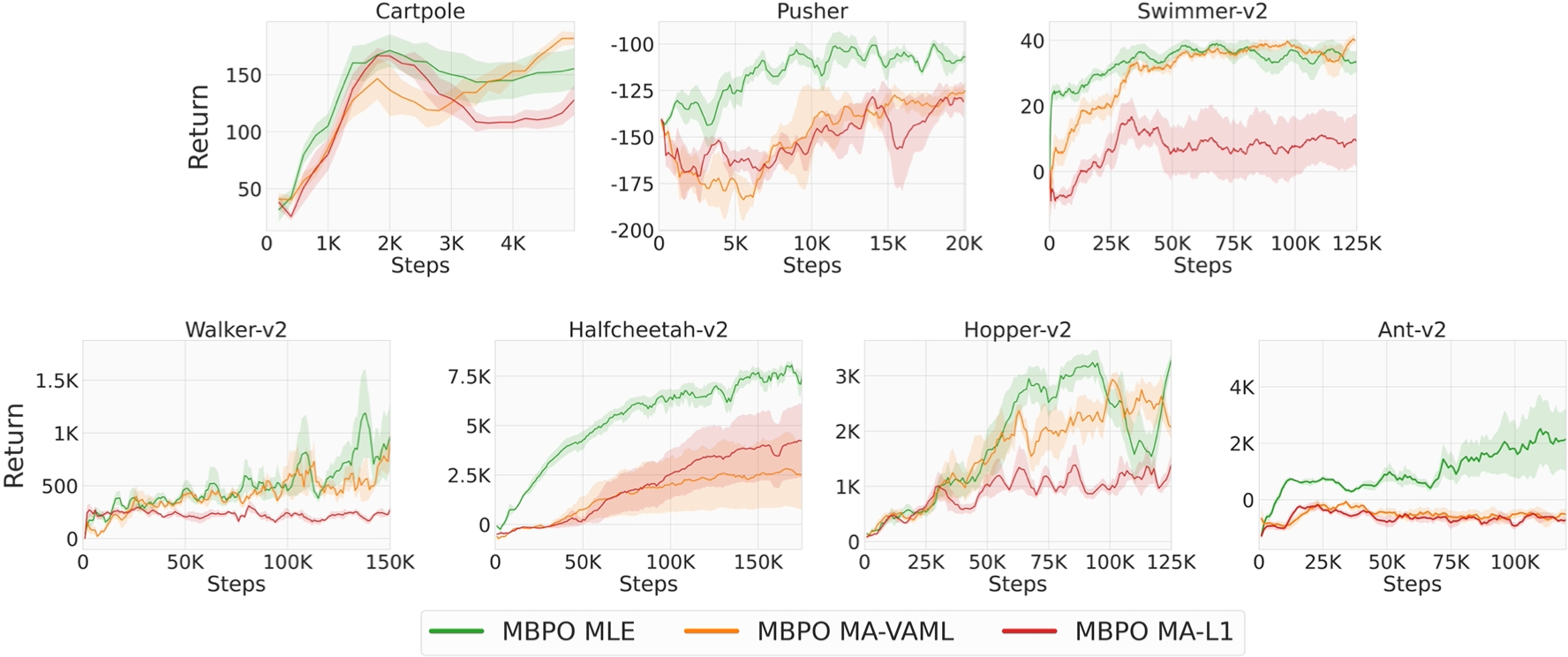}
    }
\caption{%
Evaluation on continuous control environments
for value aware methods and baselines
with MBPO~\protect\citep{janner2019trust}, 
\nmi{without tuning existing parameters, over 5 random seeds}.
\nmi{The two value-aware objectives \texttt{MBPO MA-L1} and \texttt{MBPO MA-VAML}
obtain near-matching performance with \texttt{MBPO MLE} in several environments
but under-performing  in others.
}
}
\label{fig:continuous_control_mbpo}
\end{figure*}

\subsection{Continuous Control}
\label{sec: cont_envs}

\nmi{
We select two commonly adopted dyna-style MBRL algorithms -- 
SLBO \citep{luo2018algorithmic} and MBPO \citep{janner2019trust}
as a foundation for evaluating value-aware approaches in continuous control.
For SLBO, we use their open source code\footnote{https://github.com/facebookresearch/slbo}
and for MBPO, we use the open source PyTorch implementation by 
MBRL-Lib\footnote{https://github.com/facebookresearch/mbrl-lib} \cite{mbrllib}.
In both cases, we implement two modifications -- (1) the option to swap out MLE with
value-aware losses for model-learning and (2) the option to turn on correction of stale value 
estimates as per \Cref{alg:valmbrl}. 
We tune a single new parameter -- the scaling
of the value-aware losses (which is fixed across all environments once selected
We maintain existing value for all other hyperparameters}
\hk{
in order to attribute evaluation differences solely to \Cref{alg:valmbrl} 
and value-aware losses, 
although further improvements could be achieved through tuning other hyperparameters.}

\subsection{Methods} 
\begin{figure*}[t!]
\centering
    \resizebox{0.90\textwidth}{!}{\includegraphics[width=0.98\textwidth]{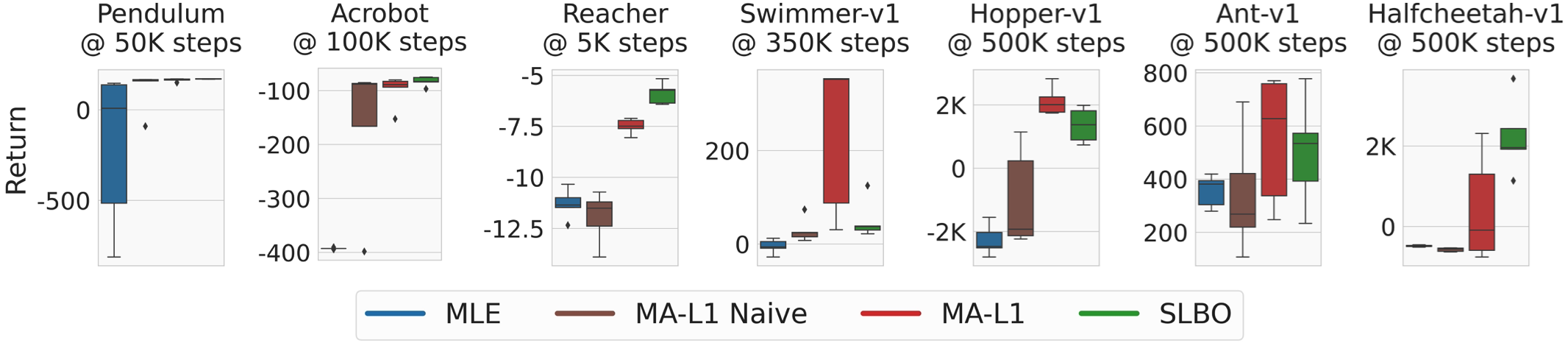}}
\caption{%
Return snapshots taken after convergence of \texttt{SLBO}, evaluated on three other
variants. \texttt{MLE} corresponds to the SLBO algorithm with just an MLE model learning objective. 
\texttt{MA-L1 Naive} corresponds to the SLBO algorithm where the model learning is objective is replaced with a value-aware objective $\calL^{U}_1$. \texttt{MA-L1} further uses \Cref{alg:valmbrl} for stale value estimate correction.
In most environments, \texttt{MA-L1} outperforms \texttt{MA-L1 Naive} and \texttt{MLE}, indicating that the stale value estimate correction of \Cref{alg:valmbrl} is the reason for improved performance.
}
\label{fig:ablation_naive}
\end{figure*}

\nmi{
For SLBO, we denote the original SLBO model-learning objective as \texttt{SLBO}. 
We denote two value-aware variants as \texttt{MA-L1} and \texttt{MA-VAML} which correspond 
to the empirical versions of 
$\calL^{U}_1$ and $\lvaml$ (IterVAML from \citep{farahmand2017value}) 
as model learning objectives respectively. Both these variants use \Cref{alg:valmbrl} i.e. the 
proposed stale value estimate correction. In \Cref{fig:ablation_naive}, we isolate
the benefits of this correction by testing the $\calL^{U}_1$ objective without \Cref{alg:valmbrl},
which we denote as \texttt{MA-L1 Naive}.
Due to the nature of the SLBO model learning objective in \citet{luo2018algorithmic},
it admits decomposition into two components -- an MLE term and a second smoothness term which minimizes 
the difference of consecutive state differences. We denote an MLE-only
baseline as \texttt{MLE}, which corresponds to keeping just the MLE term
of this objective. Intuitively, this should be a weaker baseline than SLBO as it represents
a bare bones dyna-style MBRL algorithm.
We select the same MuJoCo \cite{todorov2012mujoco} environments provided 
in the open source code by 
SLBO, shown in \Cref{fig:continuous_control_slbo}. Additionally, we show results on two OpenAI Gym \cite{brockman2016openai} environments \Pendulum{} and \Acrobot{}.
}

\nmi{For MBPO, we denote the original MBPO algorithm as \texttt{MBPO MLE}.
Two value-aware variants are obtained similar to SLBO, which we denote as 
\texttt{MBPO MA-L1} and \texttt{MBPO MA-VAML} which again correspond to the 
$\calL^{U}_1$ and $\lvaml$ 
respectively. We select the same environments provided in the open source code by MBPO, shown in \Cref{fig:continuous_control_mbpo}. 
Note that the MuJoCo environments for MBPO use the ``v2'' variants
as opposed to the ``v1'' variants in SLBO.
}

\subsection{Results} 
We present return curves for 
\nmi{SLBO variants in \Cref{fig:continuous_control_slbo},
and MBPO variants in \Cref{fig:continuous_control_mbpo}.
}
\nmi{Among the SLBO variants,}
we find that our proposed 
objective \texttt{MA-L1} outperforms \texttt{MA-VAML} on all environments 
except \Ant{} (where \texttt{MA-VAML} performs best) but still achieves
performance comparable to \texttt{SLBO}. 
\nmi{
We find a significant improvement
in performance of value-aware methods over the \texttt{MLE} baseline in 
most environments.
This is an important positive result for value-aware methods in general --
they succeed in solving continuous control tasks where MLE alone fails 
(in all environments except \Reacher{}), demonstrating that striving
for learning an accurate model (with zero MLE loss) may in fact be 
practically sub-optimal to minimizng value-aware loss functions.
}%
We also observe that on a few environments,
namely \Swimmer{} and \nma{\Hopper{}}, our \texttt{MA-L1} objective
outperforms \nma{or is competitive with} \texttt{SLBO} -- a baseline
that benefits from the smoothness \nmi{regularizer in its second term,} 
in addition to MLE.

\nmi{
Among the MBPO variants, we find that both value-aware variants \texttt{MBPO MA-L1}
and \texttt{MBPO MA-VAML}
obtain comparable but not excess return compared to \texttt{MBPO MLE}
on \CartPole{}, \Swimmertwo{}, \Pusher{}, \Walkertwo{} and \Hoppertwo{} environments -- 
indicating that they are solving the continuous control task but with lesser return.
We find that value-aware methods are not performant on \Anttwo{} and \Halfcheetahtwo{},
obtaining very low rewards (while still being positive for \Halfcheetahtwo{}).
}

\nmi{
In \Cref{fig:ablation_naive}, we find that the ablation \texttt{MA-L1 Naive} 
is outperformed by \texttt{MA-L1} in all environments except \Pendulum{}
and \Acrobot{}. The \texttt{MA-L1 Naive}
in most cases fails to exceed the performance of the \texttt{MLE} baseline,
corroborating the negative results by \citet{lovatto2020decision} and 
highlighting the importance of correcting stale value 
estimates in \Cref{alg:valmbrl}.
}

\section{Conclusion}
\nmi{
In this work, we bridge the gap in theory and practice of value-aware
model learning for model-based RL. We present a novel value-aware objective
inspired by bounding the model-advantage between an approximate and true model
given a fixed policy, demonstrating superior performance in comparison
to prior value-aware objectives in conjunction with SLBO \cite{luo2018algorithmic}.
We identify the issue of stale value estimates that hamper performance
of all value-aware methods in general if used as-is in the dyna-style MBRL 
framework. Our proposed algorithm enables successful deployment of value-aware
objectives in complex continuous control environments, representing the first
positive result in the path to bringing value-aware objectives,
well-known for their theoretical benefits, closer to practice.
We hope that these successful experimental results spur wider interest 
in value-aware model learning.
}

\bibliography{refs}
\bibliographystyle{icml2022}

\clearpage
\appendix

\section*{Appendix}

\section{Performance Difference}\label{app: mpd}
\subsection{Proof of \Cref{lemma: model-perf-diff} (Simulation Lemma)}
We restate \Cref{lemma: model-perf-diff} below, followed by the proof.

\textbf{Lemma 1.}
\textit{
    (Simulation Lemma) Let $\rM$ and $\bM$ be two different MDPs.
    Further, define 
    $\calR^\pi(s) = \EE_{a\sim\pi(\cdot|s)}[\calR(s,a)]$ 
    and $\calR^\pi_\epsilon(s) = \calR^\pi_{\rM}(s) - \calR_{\bM}^\pi(s)$.
    For any policy $\pi\in\Pi$ we have:
}
    \begin{align}
        J_\rM(\pi) &= J_\bM(\pi) 
        + \EE_{s\sim d_{\rM,\pi}}[\calR_\epsilon(s)]
        \nonumber\\&\phantom{AA}
        + \frac{1}{1-\gamma}
        \EE_{s\sim d_{\rM, \pi}}
        \EE_{\sprime\sim \Ptrue_{\rM}(\sprime\mid s,\pi)}
            \left[
                \MA^{\pi}_{\bM}(s,\sprime)
            \right] 
        \nonumber
    \end{align}

\begin{proof}
\switchDualMColor
Let $\calP_0$ be the start state distribution for both MDPs, $P^{\pi}_{\bM,t}$ be the
state distribution at time $t$ starting from $s_0\sim\calP_0$ in $\bM$, and $d_{\bM,\pi}$
denote the stationary state distribution under MDP $\bM$, policy $\pi$ and start state
$s_0\sim\calP_0$. 
We use the following slightly modified version of the definition of
value function which has a normalization of $1-\gamma$:
\begin{align*}
V^\pi_{\bM}(s_0) = (1-\gamma)\sum_{t=0}^\infty\gamma^t\EE_{a_t,s_t\sim\pi P_{\bM,t}}[\calR_{\bM}(s_t,a_t)]    
\end{align*}
Then, we have:
\begin{align*}
&J_{\bM}(\pi) - J_{\rM}(\pi)\\
&= \EE_{s_0\sim\calP_0} 
    \left[
        V^{\pi}_{\bM}(s_0) 
        - V^\pi_\rM(s_0)
    \right]\\
&= (1-\gamma)
\sum_{t=0}^\infty \gamma^t  \EE_{s_t\sim P^\pi_{\bM,t}}
\underbrace{
    \EE_{a_t\sim \pi(\cdot\mid s_t)}
    \left[
        \calR_{\bM}(s_t, a_t)
    \right]
}_{\calR_{\bM}^{\pi}(s_t)}
\nonumber\\&\phantom{AAAAAAAAAAAAAAAAAAA}
- \EE_{s_0\sim\calP_0} \left[
        V^\pi_\rM(s_0)
    \right]\\
&= (1-\gamma)
\sum_{t=0}^\infty \gamma^t  \EE_{s_t\sim P^\pi_{\bM,t}}
\left[
    \calR_{\bM}^{\pi}(s_t)
\right]
- \EE_{s_0\sim\calP_0} \left[
        V^\pi_\rM(s_0)
    \right]\\
&= \sum_{t=0}^\infty \gamma^t  \EE_{s_t\sim P^\pi_{\bM,t}}
\left[
    (1-\gamma) \calR_{\bM}^{\pi}(s_t)
    + V^\pi_\rM (s_t)
    - V^\pi_\rM (s_t) 
\right]
\nonumber\\&\phantom{AAAAAAAAAAAAAAAAAAA}
- \EE_{s_0\sim\calP_0} \left[
        V^\pi_\rM(s_0)
    \right]\\
\uatext{Cancelling the first element in the summation, and shifting the series by 1 step:}
&= \sum_{t=0}^\infty \gamma^t  \mathop{\EE}\limits_{\substack{
        s_t\sim P^\pi_{\bM,t}\\
        s_{t+1}\sim P^\pi_{\bM,t+1}
    }}
\Big[
    (1-\gamma) \calR_{\bM}^{\pi}(s_t)
    + \gamma V^\pi_\rM (s_{t+1})
    \nonumber\\&\phantom{AAAAAAAAAAAAAAAAAAA}
    - V^\pi_\rM (s_t) 
\Big]\\
\uatext{Expanding $V^\pi_\rM(s_t)$ with a one-step bellman evaluation operator:}\notag\\
&= \sum_{t=0}^\infty \gamma^t  \mathop{\EE}\limits_{\substack{
        s_t\sim P^\pi_{\bM,t}\\
        s_{t+1}\sim P^\pi_{\bM,t+1}
    }}
\Big[
    (1-\gamma) \calR_{\bM}^{\pi}(s_t)
    + \gamma V^\pi_\rM (s_{t+1})
    \nonumber\\&\phantom{AAA}
    - \left(
        (1-\gamma)\calR_{\rM}^\pi(s_t) 
        + \gamma \EE_{\sdoubleprime\sim \calP^\pi_\rM(s_t, \pi)} 
            \left[
                V^\pi_\rM(\sdoubleprime)        
            \right]
    \right)
\Big]\\
&= \sum_{t=0}^\infty \gamma^t  \mathop{\EE}\limits_{\substack{
        s_t\sim P^\pi_{\bM,t}\\
        s_{t+1}\sim P^\pi_{\bM,t+1}
    }}
\Big[
    (1-\gamma) (\calR_{\bM}^{\pi}(s_t) - \calR_{\rM}^\pi(s_t))
    \nonumber\\&\phantom{AAA}
    + \gamma V^\pi_\rM (s_{t+1})
    - \gamma \EE_{\sdoubleprime\sim \calP^\pi_\rM(s_t, \pi)} 
        \left[
            V^\pi_\rM(\sdoubleprime)        
        \right]
\Big]\\
&= \sum_{t=0}^\infty \gamma^t  \mathop{\EE}\limits_{\substack{
        s_t\sim P^\pi_{\bM,t}\\
        s_{t+1}\sim P^\pi_{\bM,t+1}
    }}
\Big[
    (1-\gamma) \calR^{\pi}_{\epsilon}(s_t)
    \nonumber\\&\phantom{AAA}
    + \gamma V^\pi_\rM (s_{t+1})
    - \gamma \EE_{\sdoubleprime\sim \calP^\pi_\rM(s_t, \pi)}
        \left[
            V^\pi_\rM(\sdoubleprime)
        \right]
\Big]\\
\uatext{Using definition of $\MA_{\rM}^\pi$:}
&= \sum_{t=0}^\infty \gamma^t  \mathop{\EE}\limits_{\substack{
        s_t\sim P^\pi_{\bM,t}\\
        s_{t+1}\sim P^\pi_{\bM,t+1}
    }}
\left[
    (1-\gamma) \calR^{\pi}_{\epsilon}(s_t)
    + \MA^\pi_\rM(s_t, s_{t+1})
\right]\\
&= \frac{1}{1-\gamma}
\EE_{s\sim d_{\bM,\pi}}
\EE_{\sprime\sim \calP_{\bM}(s,\pi)}
    \left[
        \MA^{\pi}_\rM(s,\sprime)
    \right]
\nonumber\\&\phantom{AAAAAAAAAAAA}
+ \EE_{s\sim d_{\bM, \pi}}
\left[
    \calR^\pi_\epsilon(s)
\right]
\end{align*}
\switchPrimalMColor
\end{proof}

\subsection{Proof of \Cref{cor: model-perf-diff} (Deviation Error)}
Restating \Cref{cor: model-perf-diff}:
\textbf{Corollary 2.}
\textit{
    Let $\rM$ and $\bM$ be two different MDPs.
    For any policy $\pi\in\Pi$ we have:
}
    \begin{align}
        J_\rM(\pi) &= J_\bM(\pi) 
        \nonumber\\&\phantom{A}
        + \frac{1}{1-\gamma}\EE_{s\sim d_{\rM, \pi}}
        \underbrace{
            \left[
                \calT^\pi_{\rM} V_{\bM}^\pi(s) 
                - \calT^\pi_{\bM} V_{\bM}^\pi(s)
            \right]
        }_{\text{deviation error}} 
        \label{app_eq: dev}
    \end{align}

\begin{proof}
\switchDualMColor
\begin{align*}
&J_{\bM}(\pi) - J_{\rM}(\pi)\\
&= \EE_{s_0\sim\calP_0} 
    \left[
        V^{\pi}_{\bM}(s_0) 
        - V^\pi_\rM(s_0)
    \right]\\
&=\cdots
\nonumber\\
\uatext{Proceeding similar to the previous proof upto the following line:}
&=\sum_{t=0}^\infty \gamma^t  \mathop{\EE}\limits_{\substack{
        s_t\sim P^\pi_{\bM,t}\\
        s_{t+1}\sim P^\pi_{\bM,t+1}
    }}
\Big[
    (1-\gamma) \calR_{\bM}^{\pi}(s_t)
    + \gamma V^\pi_\rM (s_{t+1})
    \nonumber\\&\phantom{AAA}
    - \left(
        (1-\gamma)\calR_{\rM}^\pi(s_t)
        + \gamma \EE_{\sdoubleprime\sim \calP^\pi_\rM(s_t, \pi)} 
            \left[
                V^\pi_\rM(\sdoubleprime)        
            \right]
    \right)
\Big] \\
&=\sum_{t=0}^\infty \gamma^t  \EE_{s_t\sim P^\pi_{\bM,t}}
\Big[
    \calT^{\pi}_{\bM}V^\pi_\rM(s_t)
    \nonumber\\&\phantom{AAA}
    - \left(
        (1-\gamma)\calR_{\rM}^\pi(s_t) 
        + \gamma \EE_{\sdoubleprime\sim \calP^\pi_\rM(s_t, \pi)} 
            \left[
                V^\pi_\rM(\sdoubleprime)        
            \right]
    \right)
\Big]\\
&=\sum_{t=0}^\infty \gamma^t  \EE_{s_t\sim P^\pi_{\bM,t}}
\left[
    \calT^{\pi}_{\bM}V^\pi_\rM(s_t)
    - \calT^{\pi}_{\rM}V^\pi_{\rM}(s_t)
\right]\\
&=\frac{1}{1-\gamma}\EE_{s\sim d_{\pi, \bM}}
\left[
    \calT^{\pi}_{\bM}V^\pi_\rM(s)
    - \calT^{\pi}_{\rM}V^\pi_{\rM}(s)
\right]
\end{align*}%
\switchPrimalMColor%
\end{proof}

\switchDualMColor
This concludes the proof of \Cref{cor: model-perf-diff}. Note that we can
further upper bound the difference in values across MDPs for a policy as
follows, which will be useful in subsequent proofs. We compute
this bound at an arbitrary start state $s_0$, and it will then hold
for any start state. Let $d_{\bM,s_0,\pi}$ be the stationary state
distribution of following policy $\pi$ in MDP $\bM$, starting
at state $s_0$.
\begin{align}
&
        V^{\pi}_{\bM}(s_0) 
        - V^\pi_\rM(s_0)
    \nonumber\\
&=\cdots\\
\uatext{Proceeding similar to the proof of \Cref{cor: model-perf-diff}, we get the following:}
&=\frac{1}{1-\gamma}\EE_{d_{\bM,s_0,\pi}}
\left[
    \calT^{\pi}_{\bM}V^\pi_\rM
    - \calT^{\pi}_{\rM}V^\pi_{\rM}
\right]\nonumber\\
&\leq \frac{1}{1-\gamma}
\linf{
    \calT^{\pi}_{\bM}V^\pi_\rM
    - \calT^{\pi}_{\rM}V^\pi_{\rM}
}\label{eq_appendix:tv_term_in_cor}\\
&\leq \frac{1}{1-\gamma}
\bigg[
    \linf{
        \calR^{\pi}_{\rM} 
        - \calR^{\pi}_{\bM}
    }
    \nonumber\\&\phantom{A}
    + \gamma\max_{s} \sum_{\sprime\in\calS} V^\pi_\rM(\sprime)
        \EE_{a\sim\pi(s)}
            \left[
                p_{\rM}(\sprime|s,a) - p_{\bM}(\sprime,a)
            \right]
\bigg]\nonumber\\
&\leq \frac{1}{1-\gamma}
\bigg[
    \epsilon_{R}
    \nonumber\\&\phantom{AAA}
    + \gamma \linf{V^\pi_\rM} 
    \underbrace{
        \max_{s}
        \max_{a}
                \norm{
                    p_{\rM}(\sprime|s,a) - p_{\bM}(\sprime,a)
                }_{1}
    }_{\epsilon_{P}}
\bigg]\nonumber\\
&\leq \frac{1}{1-\gamma}
\left[
    \epsilon_{R}
    + \gamma \linf{V^\pi_\rM}\epsilon_P
\right]\nonumber\\
&\leq \frac{1}{1-\gamma}
\left[
    \epsilon_{R}
    + \frac{\gamma\epsilon_P\Rmax}{1-\gamma}
\right]\label{eq_appendix:transfer_error_loose_bound_proof}
\end{align}
\switchPrimalMColor
\nocolorM%

\end{document}